# A Lightweight Multi-Agent Framework for Automated Concrete Barrier Design

Wanting Wang[1,*]; Xiye Ma[1,*]; Yuyang He[1]; and Ran Cao[2,†]

## Abstract

The design of reinforced concrete (RC) highway barriers is a safety-critical engineering task that requires strict compliance with regulatory provisions, such as the AASHTO LRFD Bridge Design Specifications. Current engineering practices rely largely on manual, iterative, and experience-driven procedures to satisfy complex material, geometric, and mechanical constraints. Although standalone general-purpose Large Language Models (LLMs) exhibit strong capabilities in knowledge representation and text generation, their direct application to structural engineering design remains challenging due to hallucination risks, numerical reasoning errors, and insufficient integration with physics-based analysis. To address these limitations, this study proposes a novel "generation-validation-modification" closed-loop framework for AI-assisted automated RC barrier design based on the multi-agent orchestration capability of AutoGen. Unlike standalone LLMs that directly generate design parameters from user prompts, the proposed framework integrates specialized agents for parameter generation, external mechanics-based calculation, target-interval evaluation, deviation diagnosis, and rule-based design modification. The performance of the proposed multi-agent framework (MAF) was evaluated using three barrier testing levels (TL-3, TL-4, and TL-5), including sixty RC barrier designs with different geometric configurations. Three DeepSeek models with different parameter scales, including DeepSeek (DS)-8B, DS-32B, and DS-671B, were investigated. All design results were evaluated according to Section 13 of the AASHTO LRFD Bridge Design Specifications, 10th Edition (2024). The results demonstrate that the proposed agentic framework (MAF-DS-8B) achieves a target-interval compliance rate of 98.3%, whereas the best-performing standalone LLM, DS-32B, achieves only 11.7%. These results highlight the potential of multi-agent architectures to improve the reliability, interpretability, and accessibility of AI-assisted engineering design systems for practical applications. The source code, agent prompts, and representative testing cases are publicly available at the project GitHub repository: https://github.com/MXY820/barrier-design.

## Introduction

Reinforced concrete barriers constitute a critical component of highway infrastructure, serving as the primary line of defense against vehicle encroachment and ensuring the safety of motorists, pedestrians, and adjacent structures. The design of these safety-critical elements is governed by stringent performance requirements, most notably the American Association of State Highway and Transportation Officials Load and

---

[1] Research Assistant, College of Civil Engineering, Hunan University, Changsha, 410082, China.

[2] Assistant Professor, College of Civil Engineering, Hunan University, Changsha, 410082, China.

* Equal contribution.

[†] Corresponding author: rcao@hnu.edu.cn

Resistance Factor Design (AASHTO-LRFD) Bridge Design Specifications, Section 13, which mandates rigorous evaluation of ultimate transverse resistance under vehicular impact loading [1]. Current practice in barrier design typically involves engineers performing yield-line analysis calculations either by hand or through bespoke spreadsheet implementations of AASHTO-LRFD provisions. This methodology requires not only familiarity with complex limit state formulations but also deep expertise in reinforced concrete designs, material nonlinearity, and failure mode identification. Consequently, the design process is highly iterative and trial-and-error driven: engineers must propose initial geometries and reinforcement layouts, evaluate the resulting ultimate resistance capacity ($R_w$), compare it against required transverse impact forces ($F_t$) for specified test levels, and manually adjust parameters until compliance is achieved. This manual loop highlights a compelling need for computational methodologies capable of intelligently navigating the barrier design space while preserving the physical rigor mandated by governing specifications.

The past decade has witnessed transformative advances in artificial intelligence, with Large Language Models (LLMs) emerging as particularly versatile tools capable of processing natural language specifications and generating structured outputs across diverse domains [2–5]. In engineering contexts, researchers have begun exploring the application of LLMs to tasks ranging from code generation and technical documentation synthesis to conceptual design exploration and parametric modeling [6–8]. In the specific domain of the architecture, engineering, and construction (AEC) industry, preliminary investigations have demonstrated the potential for LLMs to assist in automated code compliance checking, preliminary sizing of structural elements, and conceptual structural analysis [9–15]. However, these applications remain largely exploratory, and their translation to safety-critical design tasks, where analytical errors carry profound consequences, has been appropriately cautious.

The direct application of LLMs to structural engineering design tasks encounters several fundamental obstacles that preclude their deployment as autonomous design agents. Foremost among these is the challenge of physical grounding; trained primarily on disembodied text, LLMs lack an intrinsic understanding of mechanical principles, material behavior, and spatial-geometric constraints [16,17]. Moreover, the stochastic nature of LLM generation, while highly advantageous for creative ideation, presents unacceptable hallucination risks and reliability concerns for safety-critical infrastructure [18]. To bridge this gap, recent pioneering studies in broader engineering disciplines have demonstrated that integrating domain-specific physical principles with AI agents can successfully guide generative models through complex engineering spaces, such as automating advanced alloy discovery [19].

Meanwhile, a significant advancement in AI system architecture has been the development of multi-agent collaborative frameworks [20-25], exemplified by platforms such as Microsoft's AutoGen [26], which enables the orchestration of multiple specialized LLM instances collaborating toward shared objectives. Unlike monolithic LLM deployments where a single model bears the sole responsibility for both reasoning and output generation, agentic frameworks distribute cognitive labor across distinct, specialized roles, such as planners, critics, executors, and optimizers, that communicate through structured conversational protocols [21,23]. This multi-agent architectural paradigm has demonstrated particular efficacy in complex problem-solving domains requiring iterative refinement, robust error detection, and strict constraint satisfaction [20,22,26]. This offers a highly promising pathway to overcome the limitations of single-agent systems in structural engineering design.

By leveraging the agentic orchestration capabilities of the AutoGen framework, this study proposed a "generation-validation-modification" closed-loop architecture to automate the design of reinforced concrete barriers. As illustrated in Figure 1, the workflow is organized into five coupled modules that transform unstructured user intent into validated engineering outputs through iterative validation and rule-based modification.

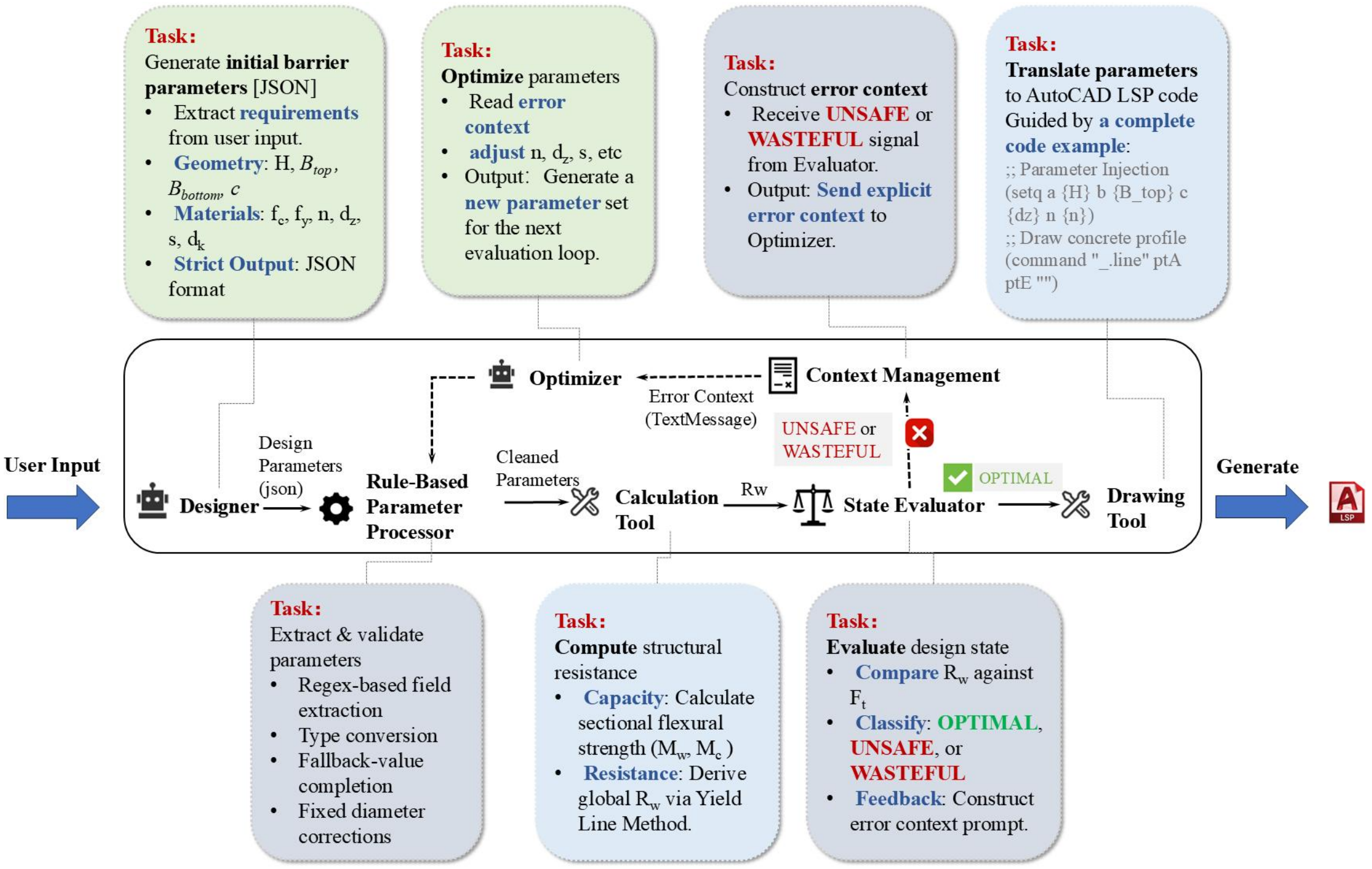

Figure 1. Proposed framework of the intelligent barrier design

# Multi-Agent Design Framework for Concrete Barriers

## *Rule-based Parameter Processor*

The workflow begins with user-provided natural language specifications, including safety requirements, test level and geometric constraints. A Designer Agent interprets these inputs and produces an initial design proposal in the form of a structured JSON parameter set, determining key geometric variables ($H$, $B_{top}$, $B_{bottom}$) and material variables ($f_c$, $f_y$, $n$, $d_z$, $s$, $d_k$).

Because LLM-generated outputs are inherently stochastic, the proposed parameter set is subsequently processed by a Rule-based Parameter Processor, which performs deterministic field extraction, type conversion, and rule-based validation on the raw model response. Specifically, predefined parameter fields are extracted using regular-expression-based mapping, converted to their required data types, and stored in a fixed parameter dictionary. Rather than relying on a formal JSON Schema validator, the current implementation defines the expected fields, data types, and allowable ranges through this predefined dictionary, as summarized in Appendix Table A2.

It is important to distinguish these deterministic preprocessing operations from the design-level parameter adjustment performed later by the Optimizer Agent. The Rule-based Parameter Processor only standardizes and sanitizes the LLM output to ensure compatibility with downstream computation, whereas the Optimizer Agent modifies the design variables based on mechanics evaluation and code compliance. The resulting validated parameter dictionary is then passed to the external mechanics calculation tool for structural analysis. Table 1 summarizes the definitions, symbols, and units of the abbreviated design parameters used throughout the framework.

Table 1. Design Parameter Definitions

| Symbol | Description | Unit |
|---|---|---|
| grade | Barrier performance level (e.g., TL-3, TL-4) | - |
| $H$ | Total height of the barrier | mm |
| $B_{top}$ | Top width of the single-slope section | mm |
| $B_{bottom}$ | Bottom width of the single-slope section | mm |
| $c$ | Concrete cover thickness | mm |
| $f_c$ | Concrete compressive strength | Mpa |
| $f_{y1}$ | Yield strength of stirrups | Mpa |
| $f_{y2}$ | Yield strength of longitudinal bars | Mpa |
| $s$ | Spacing of stirrups | mm |
| $n$ | Total number of longitudinal bars | - |
| $d_z$ | Diameter of longitudinal bars | mm |
| $d_k$ | Diameter of stirrups | mm |

## *Mechanics Calculation and State Evaluation*

Structural performance is evaluated based on the ultimate transverse resistance $R_w$, which is computed using yield line theory (Equations (1) and (2)) in accordance with AASHTO LRFD Bridge Design Specifications [1].

$$R_w=\left(\frac{2}{2L_c-L_t}\right)\left(M_w+\frac{M_c L_c^2}{H}\right) \tag{1}$$

$$L_c=\frac{L_t}{2}+\sqrt{\left(\frac{L_t}{2}\right)^2+H\left(\frac{M_w}{M_c}\right)} \tag{2}$$

where： $H$ = height of wall (m)

$L_c$ = critical length of yield line failure pattern (m)

$L_t$ = longitudinal length of distribution of impact force (m)

$R_w$ = total transverse resistance of the railing (kN)

$M_c$ = flexural resistance of cantilevered walls about an axis parallel to the longitudinal axis of the bridge (kN)

$M_w$ = flexural resistance of the wall about its vertical axis (kN·m)

The computed resistance $R_w$ is compared against the required impact force $F_t$. The design state is classified as:

- **UNSAFE**: $R_w < F_t$
- **WASTEFUL**: $R_w > \beta F_t$(e.g., $\beta = 1.6$)
- **OPTIMAL**: $\alpha F_t < R_w < \beta F_t$ , within a prescribed tolerance (e.g., $\alpha = 1.4, \beta = 1.6$)

This classification provides quantitative feedback for iterative modification and does not by itself establish global engineering optimality or comprehensive compliance with every AASHTO LRFD provision.

### *Context Management and Closed-Loop Modification*

For non-optimal states (unsafe or wasteful), the system constructs a structured error context based on the deviation between $R_w$ and the nearest target bound. This context is passed to an Optimizer Agent, which performs targeted parameter updates before recalculation.

The agent applies rule-based and mechanics-informed adjustments, such as:

- Increasing reinforcement or section dimensions for below-target designs.
- Reducing reinforcement or section dimensions for above-target designs.

The updated parameters are revalidated and re-evaluated, forming a closed-loop process. The iteration terminates when the calculated resistance falls within the target interval or when the predefined maximum number of cycles is reached, in which case the best available solution will be returned.

### *Automated Drafting and Output Generation*

Once a within-target design is obtained, the system generates a parametric AutoLISP script [27] that encodes the full geometry and reinforcement layout. This script can be directly executed in a CAD environment to produce standardized 2D drawings.

The generated geometry not only supports engineering documentation but also provides structured data (e.g., nodal coordinates) for downstream applications such as numerical simulation and 3D modeling. A detailed example of the design process is shown in Figure 2.

## Experimental Design

To comparatively evaluate the target-interval adherence of the proposed multi-agent collaborative framework (MAF) and standalone LLM configurations, a systematic experiment was conducted. This section describes the selected models, structural design scenarios, controlled repeatability experiment, and performance metrics.

### *Model Selection and Configuration*

The experimental matrix evaluates three distinct LLMs representing various scales of computational capacity: DeepSeek-8B, DeepSeek-32B, and DeepSeek-671B [28,29]. These standalone models serve as baselines for comparison with MAF-DS-8B and MAF-DS-32B.

The comparison evaluates performance differences associated with integrating the selected models into the complete multi-agent workflow; it is not a component-wise ablation study.

### *Concrete Barrier Design Scenarios*

To ensure a comprehensive assessment across diverse impact demands, three testing levels of the barriers were selected in accordance with MASH (2016) and AASHTO LRFD (2024): TL-3, TL-4, and TL-5.

This study focuses on the single-slope barrier profile, a standard geometry in contemporary highway infrastructure. For each test level, 20 unique design cases were generated by varying the barrier height within specified ranges (see Table 2).

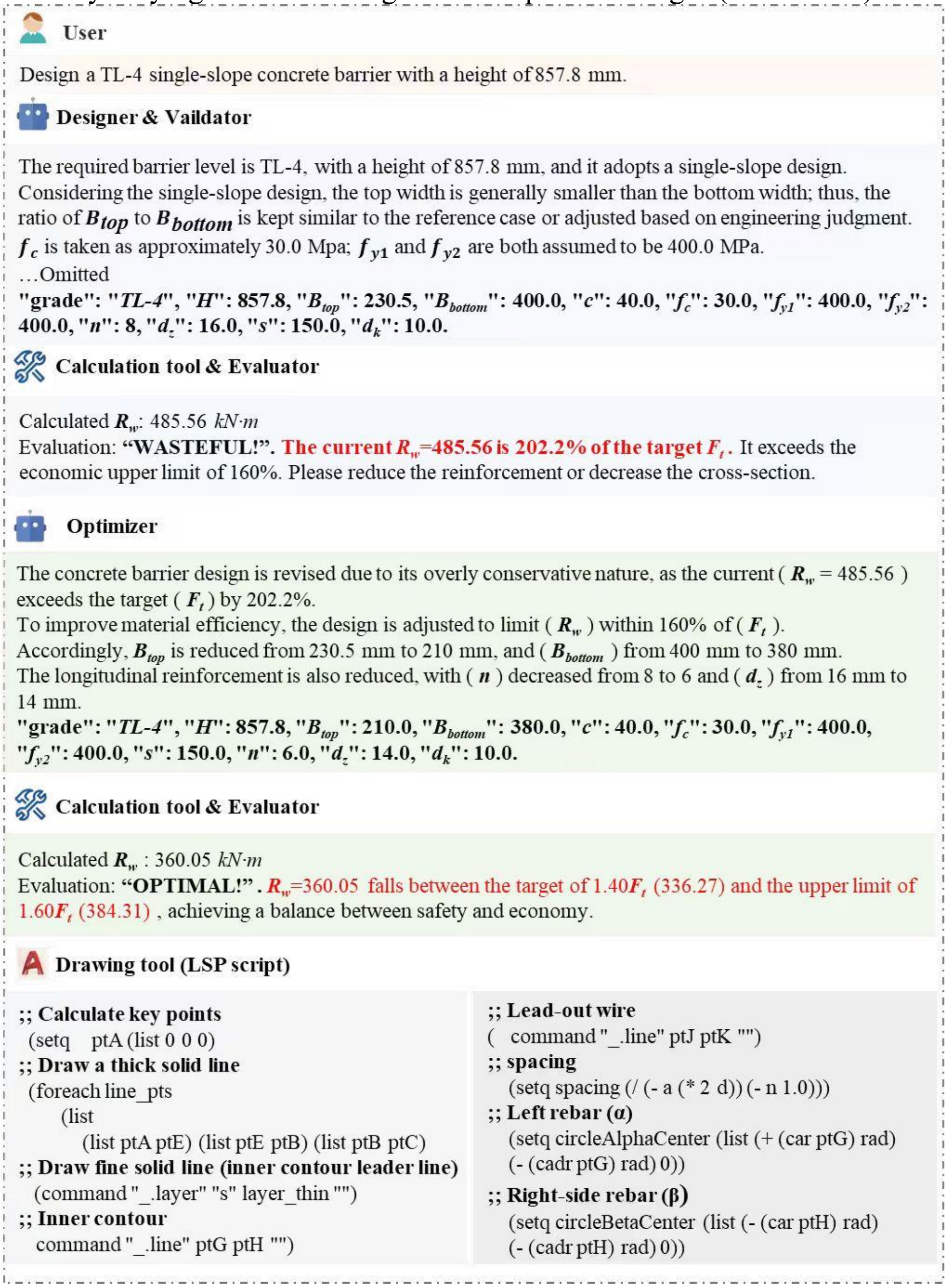

Figure 2. Example of the automated design process using the multi-agent collaboration.

Table 2. Key parameters for the experimental design

| | Parameters |
|---|---|
| Large Language Model | DeepSeek-8B, DeepSeek-32B, DeepSeek-671B |
| Barrier design levels | TL-3, TL-4, and TL-5 |
| Number of design cases for each testing level | 20 |
| Barrier shape | Single-slope |
| Barrier height | TL-3: 685.80 mm- 780.80 mm<br>TL-4: 812.80 mm- 907.80 mm<br>TL-5: 1066.80 mm-1161.80 mm |
| Controlled repeatability condition | TL-4; $H$ = 860.30 mm; $F_t$ = 240.19 kN |
| Repeated runs per configuration | 50 (standalone DeepSeek-8B and MAF-DS-8B) |

## *Controlled Repeatability Experiment*

To quantify uncertainty and repeatability independently of changes in barrier geometry, an additional controlled experiment was conducted for a TL-4 single-slope concrete barrier with a fixed height of 860.30 mm. The transverse design force was $F_t = 240.19$ kN, giving a study-defined target interval of 336.27–384.31 kN. Standalone DeepSeek-8B and MAF-DS-8B were each independently executed 50 times using the same external design task and target-interval criterion. Invalid outputs were retained in the total number of trials and counted as noncompliant. Wilson [31] 95% confidence intervals were calculated for compliance rates.

## *Target Design Criteria*

As detailed in Appendix Table A1, a review of nine actual design drawings from NCHRP RR 1109 [32] gives an average nominal-resistance-to-design-force ratio of 1.49. Allowing a practical fluctuation of approximately 10%, this study adopts $1.4F_t$ to $1.6F_t$ as a study-defined heuristic target interval for comparative evaluation. The interval is not a universally prescribed optimum in the AASHTO LRFD specifications and should not be interpreted as proof of minimum cost, minimum material use, or global engineering optimality. The framework iteratively modifies barrier parameters to move the calculated resistance toward this target interval.

## *Evaluation Metrics*

The target-interval compliance rate is defined as the proportion of experimental trials in which the calculated ultimate transverse resistance $R_{w,i}$ falls within the study-defined target interval (Equation 3).

$$Compliance\ Rate = \frac{1}{N}\sum_{i=1}^{N} I\left(L \le R_{w,i} \le U\right) \times 100\% \qquad (3)$$

where $R_{w,i}$ represents the calculated ultimate transverse resistance in the $i^{th}$ trial; $L$ and $U$ denote the lower and upper bounds of the study-defined target interval, respectively; $N$ is the total number of experimental trials; and $I$ is an indicator function that equals 1 when the calculated resistance falls within the target interval and 0 otherwise.

To quantitatively characterize the extent to which the calculated resistance deviates from the study-defined target interval, an interval-violation metric is introduced. For the $i^{th}$ trial, the interval violation, $e_i$, is defined as:

$$e_i = \begin{cases} 0, & L \le R_{w,i} \le U \\ L - R_{w,i}, & R_{w,i} < L \\ R_{w,i} - U, & R_{w,i} > U \end{cases} \tag{4}$$

In essence, the interval violation $e_i$ represents the shortest distance between the calculated resistance $R_{w,i}$ and the study-defined target interval $[L, U]$. If $R_{w,i}$ lies within the target interval, $e_i$ is equal to zero. If $R_{w,i}$ falls below the lower bound $L$, $e_i$ is calculated as the distance ($L$-$R_{w,i}$); if $R_{w,i}$ exceeds the upper bound $U$, $e_i$ is calculated as the distance ($R_{w,i}$- $U$). Therefore, a larger value of $e_i$ indicates a greater deviation of the calculated resistance from the target interval.

To assess the overall deviation of the generated design values, the Mean Squared Error (MSE) over the entire dataset is computed as:

$$MSE_{interval} = \frac{1}{N}\sum_{i=1}^{N}(e_i)^2 \tag{5}$$

# Results and Discussion

## *Comparative Performance of Standalone LLMs*

The performance of standalone DeepSeek (DS) models in the direct design of TL-4 concrete barriers is illustrated in Figure 3. The generated resistance values frequently deviate from the study-defined target interval of ($1.4F_t$–$1.6F_t$).

As shown in Figure 3a, the DS-8B produces large output variability and several extreme outliers, including both below-target and markedly above-target resistance values. DS-32B and DS-671B show smaller extreme deviations in some cases, but most outputs still fall outside the target interval. These observations indicate limited target-interval adherence under the evaluated standalone settings.

Quantitatively, the overall target-interval compliance rates for DS-8B, DS-32B, and DS-671B are merely 6.7%, 11.7%, and 8.3%, respectively (Table 3). Figure 4 presents representative unsuccessful outputs from the standalone models. Under the selected task and prompting conditions, increasing model scale alone did not consistently improve target-interval adherence. Results for TL-3 and TL-5 are provided in the Appendix.

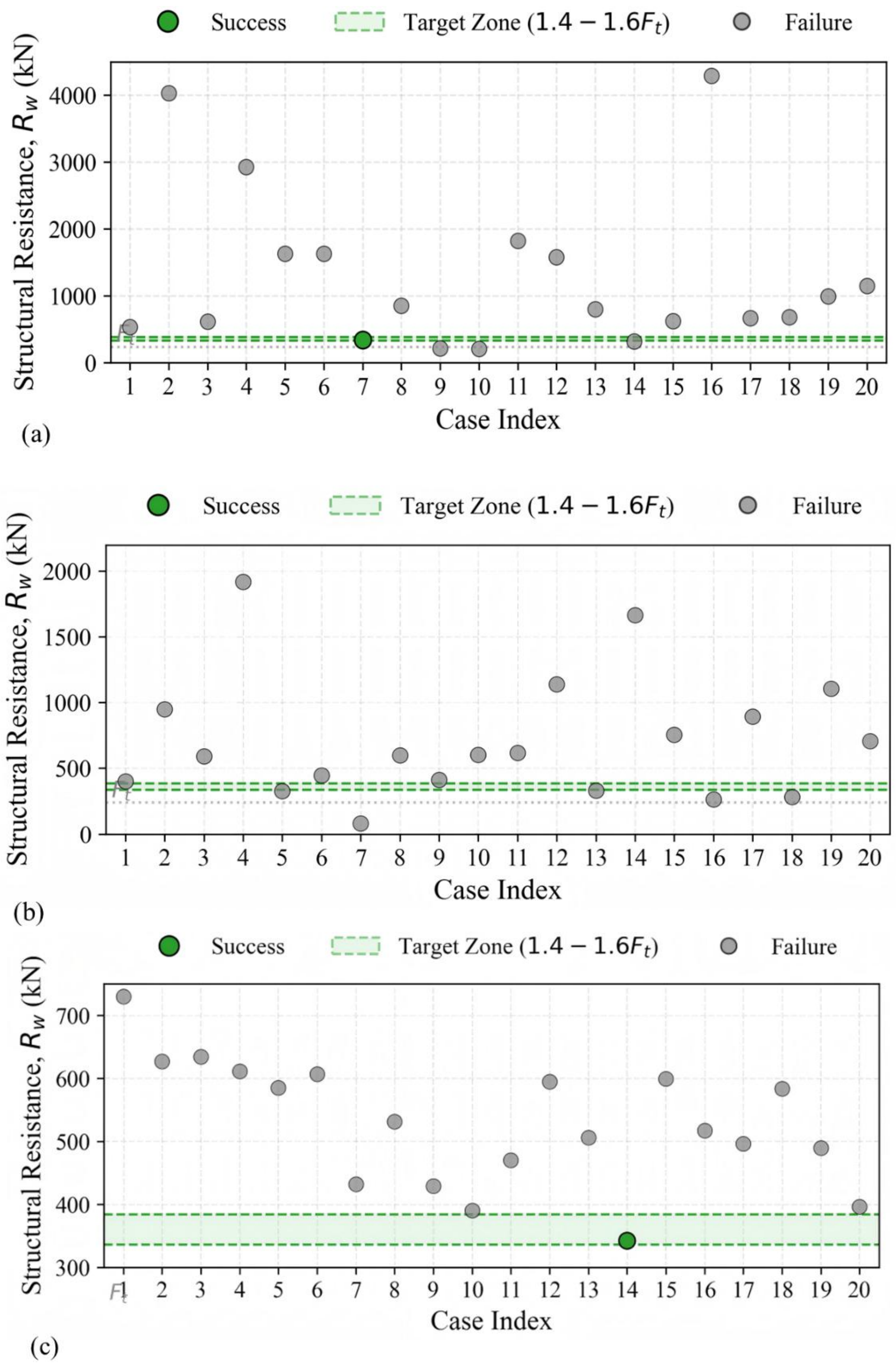

Figure 3. Target-interval compliance results for TL- 4 barriers using general large language models: (a) DS-8B; (b) DS-32B; (c) DS-671B

Table 3. Comparison of Target-Interval Compliance Rates Across Different Protection Levels

| Experimental Group | TL-3 | TL-4 | TL-5 | Compliance Rate |
|---|---|---|---|---|
| DeepSeek 8B | 0% | 5% | 15% | 6.7% |
| DeepSeek 32B | 10% | 0% | 25% | 11.7% |
| DeepSeek 671B | 5% | 5% | 15% | 8.3% |
| MAF-DS-8B | 100% | 100% | 98.3% | 98.3% |
| MAF-DS-32B | 80% | 90% | 95% | 88.3% |

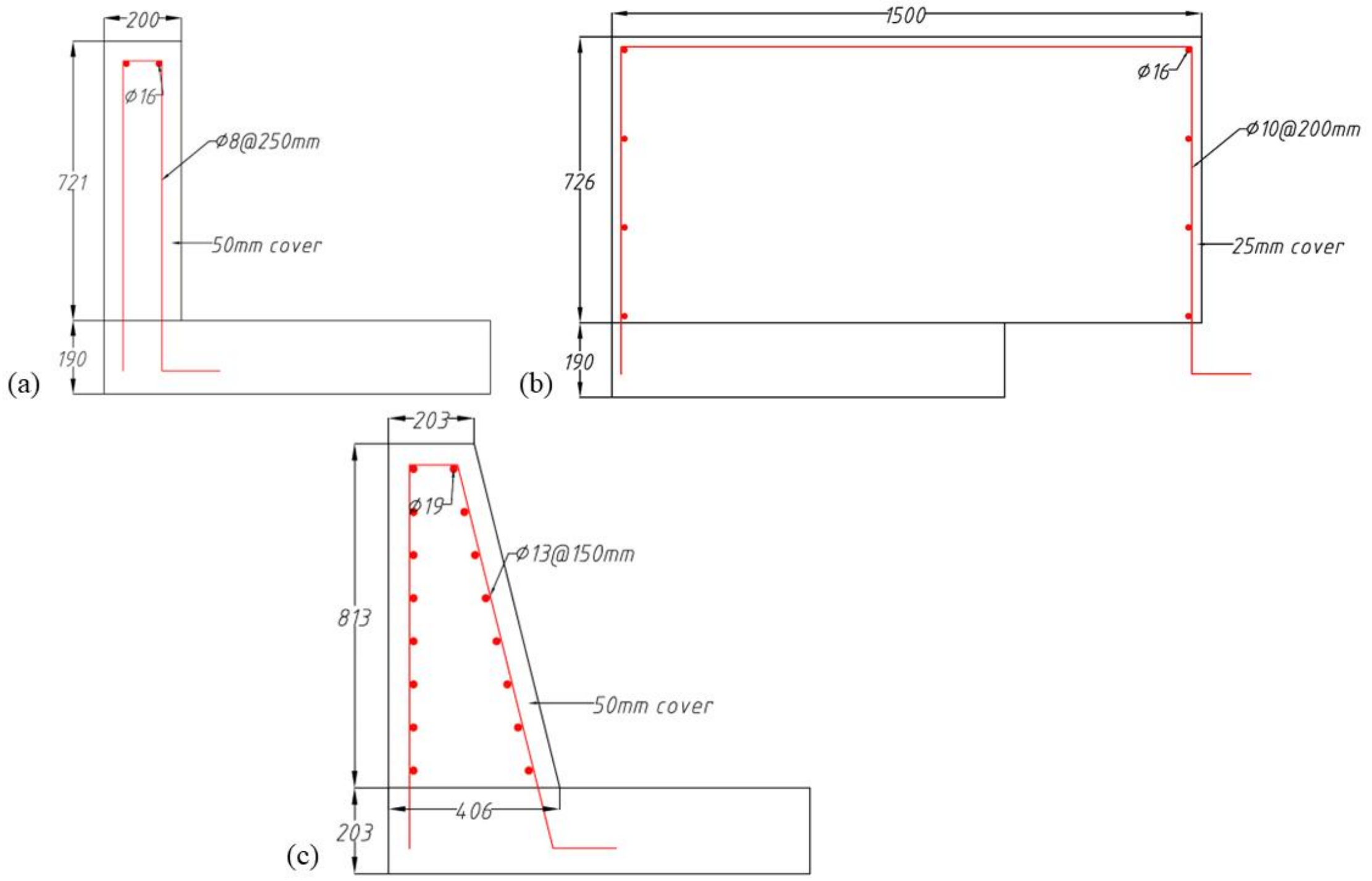

Figure 4. Representative standalone-LLM outputs: (a) below-target design; (b) geometrically inconsistent design; (c) above-target design.

## *Efficacy of the Multi-Agent Collaborative Framework*

As shown in Figure 5 and Table 3, the proposed MAF configurations achieve substantially higher target-interval compliance rates than the corresponding standalone LLM configurations in the original multi-condition evaluation. The overall compliance rates are 98.3% for MAF-DS-8B and 88.3% for MAF-DS-32B, compared with 6.7%, 11.7%, and 8.3% for standalone DS-8B, DS-32B, and DS-671B, respectively. These results support improved target-interval adherence, but do not by themselves demonstrate global engineering optimality.

Figure 7b and Table 6 summarize the mean squared interval violation, which measures the magnitude of resistance violations outside the target interval rather than conventional prediction error. The lower values obtained by the MAF configurations are associated with the complete closed-loop workflow integrating parameter generation, external mechanics calculation, target-interval evaluation, error feedback, and rule-based modification.

Under the evaluated settings, MAF-DS-8B achieved a higher target-interval compliance rate than MAF-DS-32B and the standalone DeepSeek-671B configuration. This task-specific observation suggests that structured agent collaboration may enable a lightweight model to achieve competitive performance.

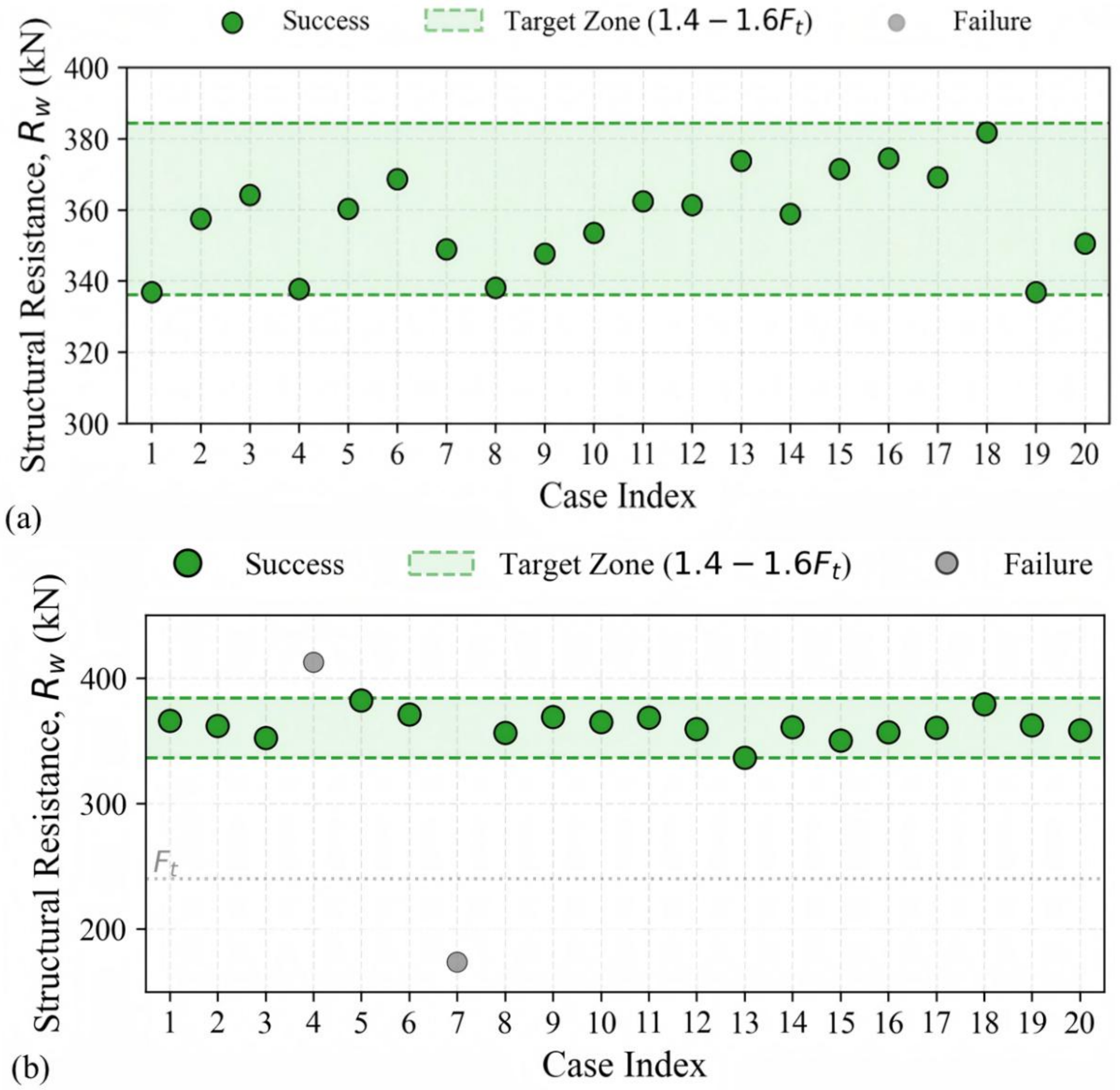

Figure 5. Target-interval compliance results for TL-4 barriers using the multi-agent framework: (a) MAF-DS-8B; (b) MAF-DS-32B.

To provide uncertainty estimates and a controlled baseline comparison, we independently executed standalone DeepSeek-8B and MAF-DS-8B fifty times each under a fixed TL-4 condition. The statistical results are summarized in Table 4. As shown in the table, when using the same foundation model (DS-8B), MAF produced substantially more stable and compliant design results than the standalone LLM.

Table 4. Statistical comparison between standalone DeepSeek-8B and MAF-DS-8B under a fixed TL-4 design condition.

| Metric | Standalone DeepSeek-8B | MAF-DS-8B |
|---|---|---|
| Number of trials | 50 | 50 |
| Within-target outputs | 2 | 49 |
| Out-of-target outputs | 48 | 1 |
| Target-interval compliance rate | 4.0% | 98.0% |
| Wilson 95% CI | 1.1%-13.5% | 89.5%-99.6% |
| Mean | 1499.15 kN | 360.98 kN |
| Sample Variance | 1621.82 kN | 14.11 kN |
| $MSE_{interval}$ | $3.82\times10^6$ $kN^2$ | 9.35 $kN^2$ |

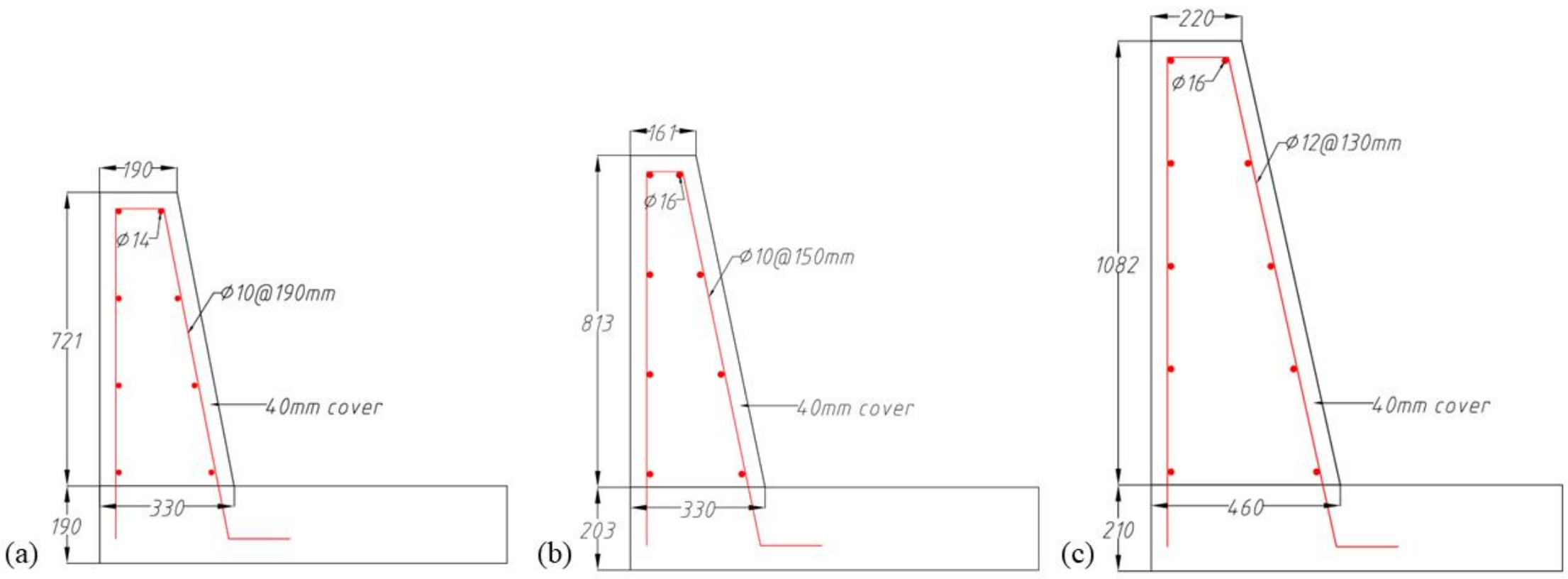

Figure 6. Representative concrete barrier designs generated by the multi-agent framework: (a) TL-3; (b) TL-4; (c) TL-5.

Table 5. Design parameters and verification results for the representative barriers shown in Figure 6.

| Parameter | Unit | TL-3 | TL-4 | TL-5 |
|---|---|---|---|---|
| $F_t$ | kN | 240.19 | 240.19 | 551.55 |
| $f_c$ | MPa | 30 | 30 | 30 |
| Reinforcement yield strength | MPa | 420 | 400 | 400 |
| Calculated $R_w$ | kN | 348.81 | 366.31 | 840.56 |
| Target interval check (manual) | - | Yes | Yes | Yes |

The resistance values in Table 5 were independently recalculated using a separate verification script based on Equations (1) and (2). The checks confirm numerical consistency and adherence to the study-defined resistance interval.

Figure 6 and Table 5 present representative design examples generated by MAF-DS-8B. Manual evaluation confirms that these designs satisfy the relevant code requirements while exhibiting reasonable structural configurations that closely resemble the design drawings produced by professional engineers.

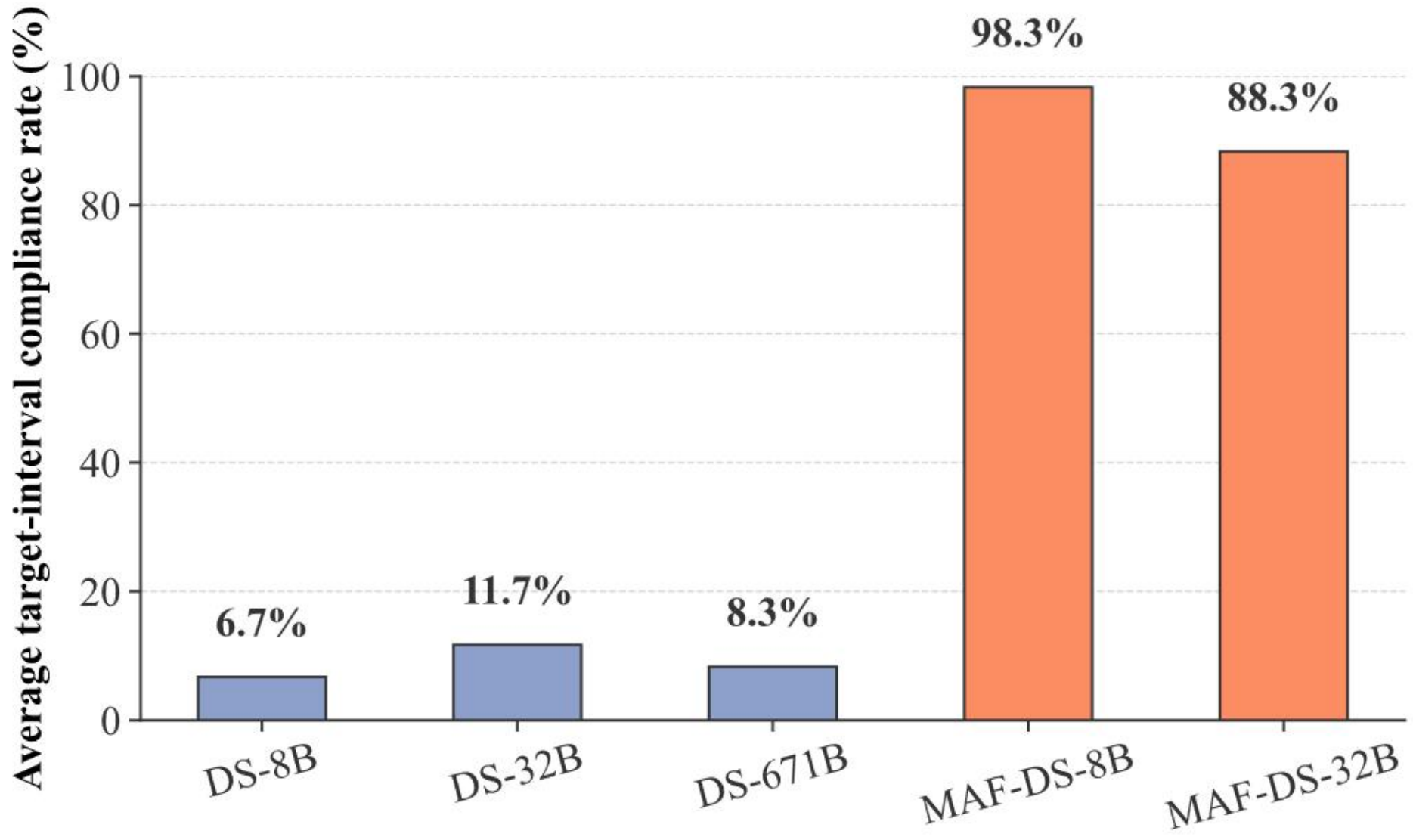

(a)

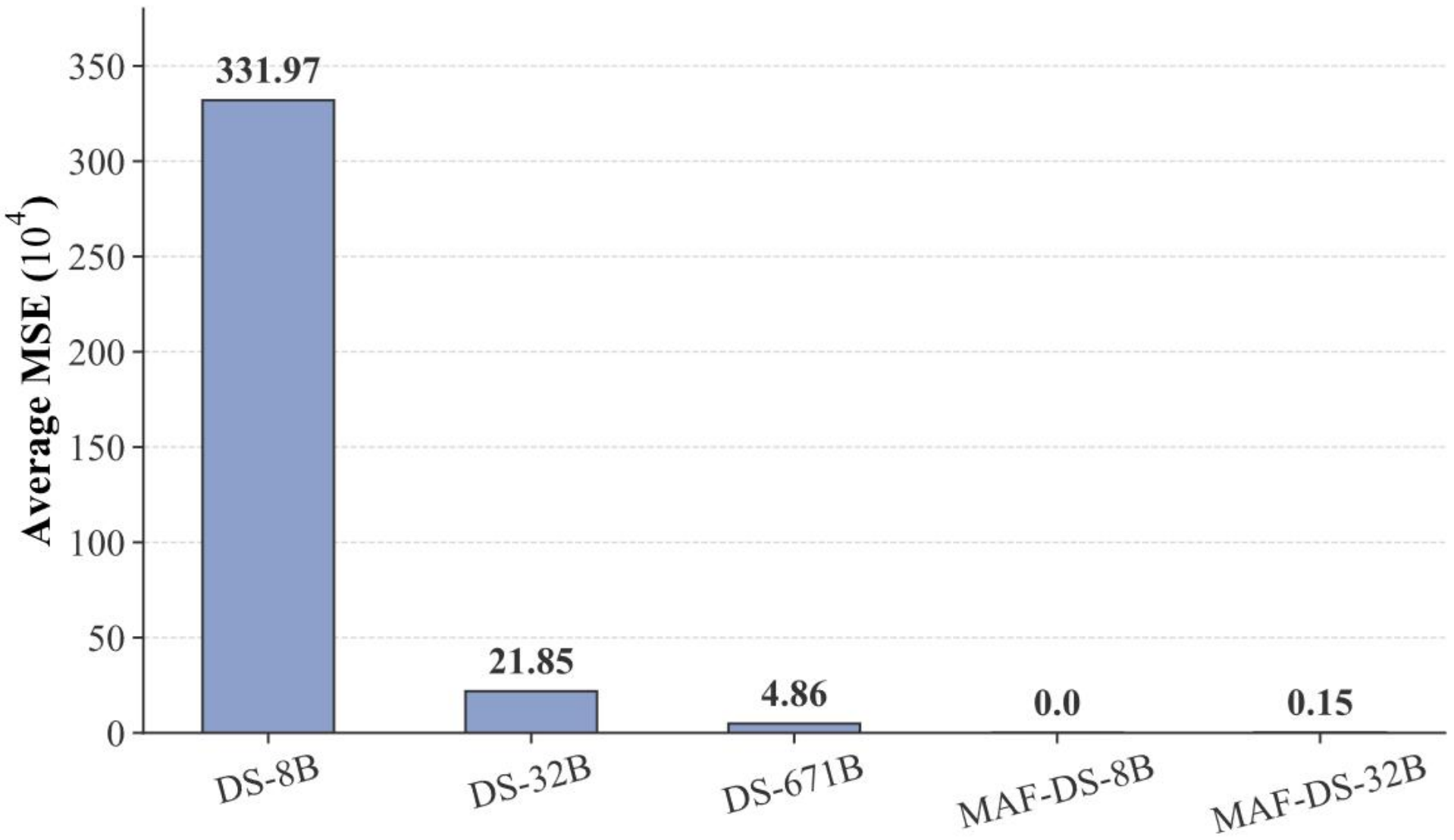

(b)

Figure 7. Quantitative comparison of the generated design results using different models: (a) Average target-interval compliance rate; (b) Average MSE.

Table 6. Comparison of mean squared interval violation for concrete barriers using different LLM configurations.

| Experimental Group | TL-3 ($10^4$) | TL-4 ($10^4$) | TL-5 ($10^4$) | Average MSE ($10^4$) |
|---|---|---|---|---|
| DeepSeek 671B | 2.91 | 3.01 | 8.65 | 4.86 |
| DeepSeek 32B | 25.31 | 30.89 | 9.35 | 21.85 |
| DeepSeek 8B | 355.44 | 216.63 | 423.83 | 331.97 |
| MAF-DS-32B | 0.19 | 0.14 | 0.12 | 0.15 |
| MAF-DS-8B | 0 | 0 | 0 | 0 |

# Limitations

The present study has several limitations that highlight key directions for future research. First, while evaluation metrics such as the target-interval compliance rate and deviation magnitude assess adherence to prescribed design drawing specifications, they do not directly quantify global engineering optimality, such as material consumption or construction cost. Second, the current comparison between standalone LLMs and the proposed Multi-Agent Framework (MAF) workflow represents a system-level baseline evaluation rather than a component-wise ablation study. Consequently, future work will separately isolate and evaluate the individual effects of deterministic validation, external mechanics calculations, state evaluation, and rule-based modifications. Third, more detailed statistical and reliability analyses are needed to rigorously evaluate agent performance and account for operational uncertainties. Finally, our model size comparisons were limited to the DeepSeek series; evaluating a broader spectrum of foundation models using the proposed benchmark remains an important subject for future investigation.

# Conclusion

This paper presented a novel multi-agent collaborative framework tailored for the automated design of concrete barriers. By structuring the design process into a modular agentic architecture integrated with deterministic validation layers and rule-based modification routines, the framework successfully bridges the gap between generative artificial intelligence and safety-critical structural engineering.

The empirical findings of this study yield several critical insights and contributions to the field of AI-assisted engineering:

- The proposed agentic methodology achieved target-interval compliance rates exceeding 98%. This represents a transformative advancement over standalone, general-purpose LLMs, which inherently lack the domain-specific constraints required for structural engineering tasks.

- Operating within the proposed framework: an 8B-parameter lightweight foundation model consistently outperformed unconstrained 671B-parameter flagship models. This demonstrates that specialized, constrained architectures can achieve superior localized efficacy for specific design routine tasks.

- By successfully orchestrating physical guardrails, structured error feedback, and heuristic parameter refinement, this work establishes a replicable template for integrating AI into highly regulated, code-compliant structural design domains.

While the current framework demonstrates robust capabilities in generating code-compliant structural designs, future research will focus on extending this agentic pipeline toward end-to-end engineering workflows. Specifically, the design outputs produced by the current system will serve as inputs to a downstream agentic framework that automatically generates high-fidelity simulation models in commercial finite element analysis (FEA) platforms, such as ANSYS and LS-DYNA. This extension will establish a seamless connection between conceptual design generation and physics-based validation, advancing toward fully autonomous, closed-loop structural engineering systems.

## Appendix

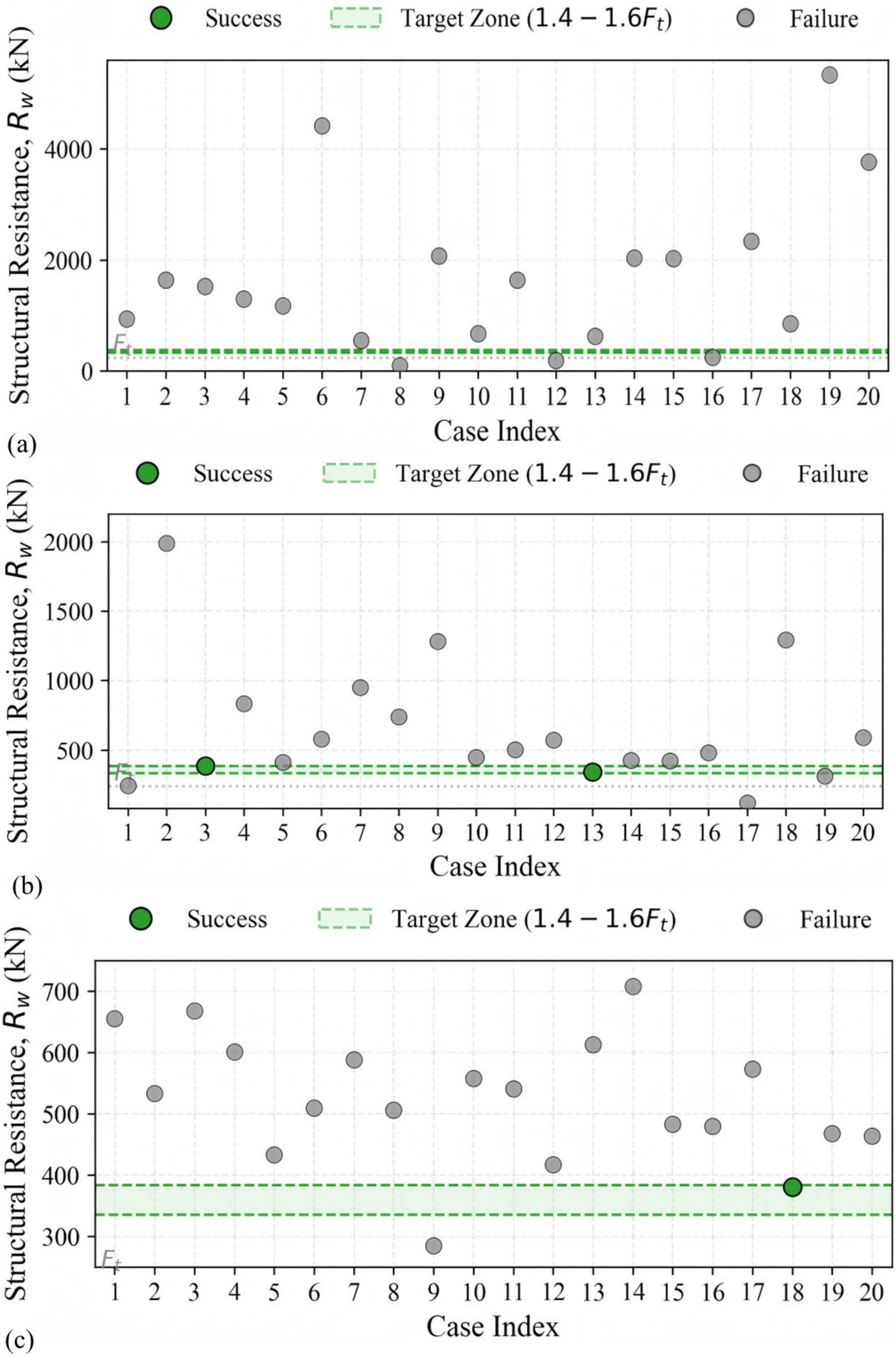

Figure A1. Target-interval compliance results of TL- 3 barriers using general large language models: (a) DS-8B; (b) DS-32B; (c) DS-671B

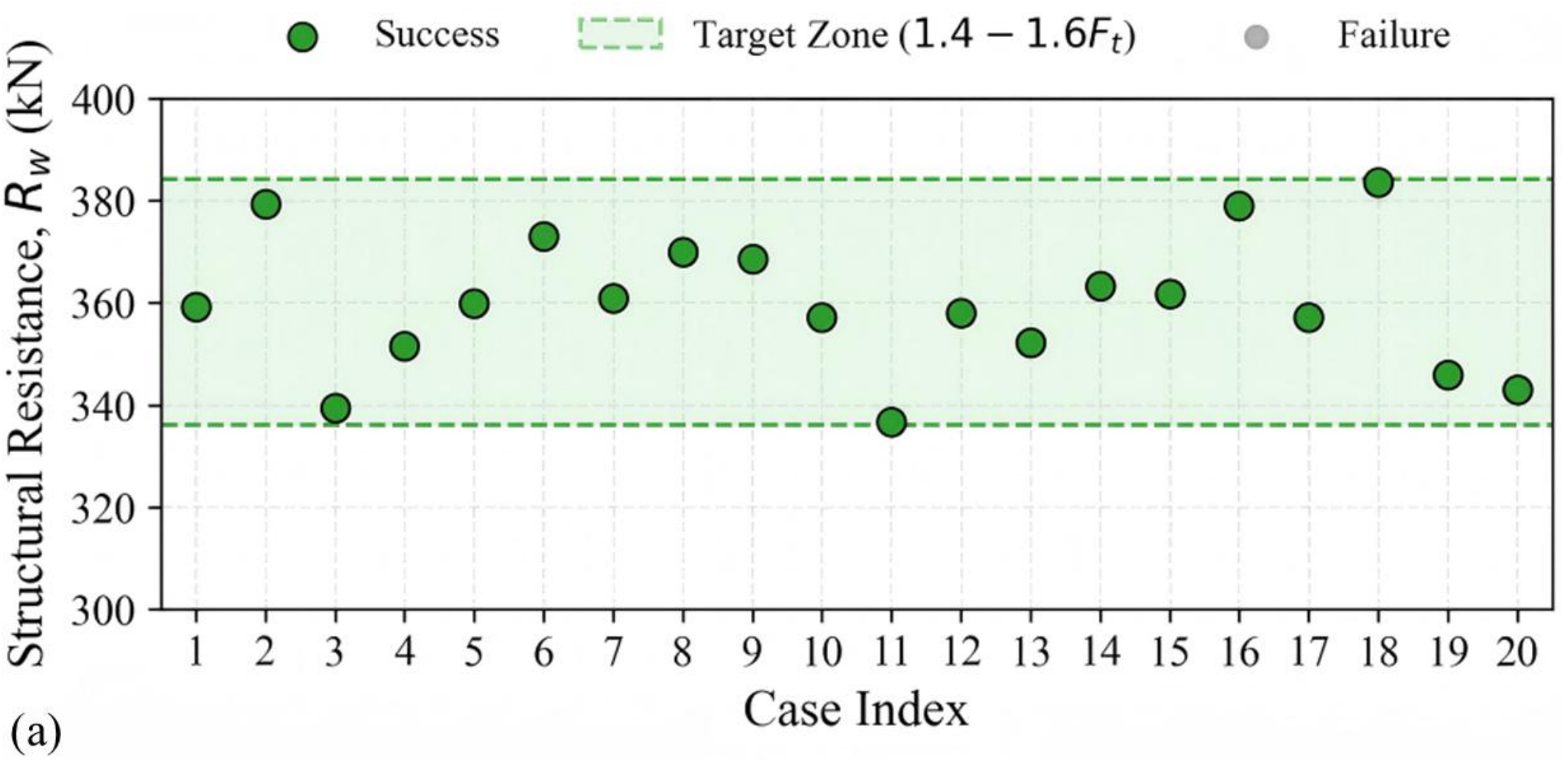

(a)

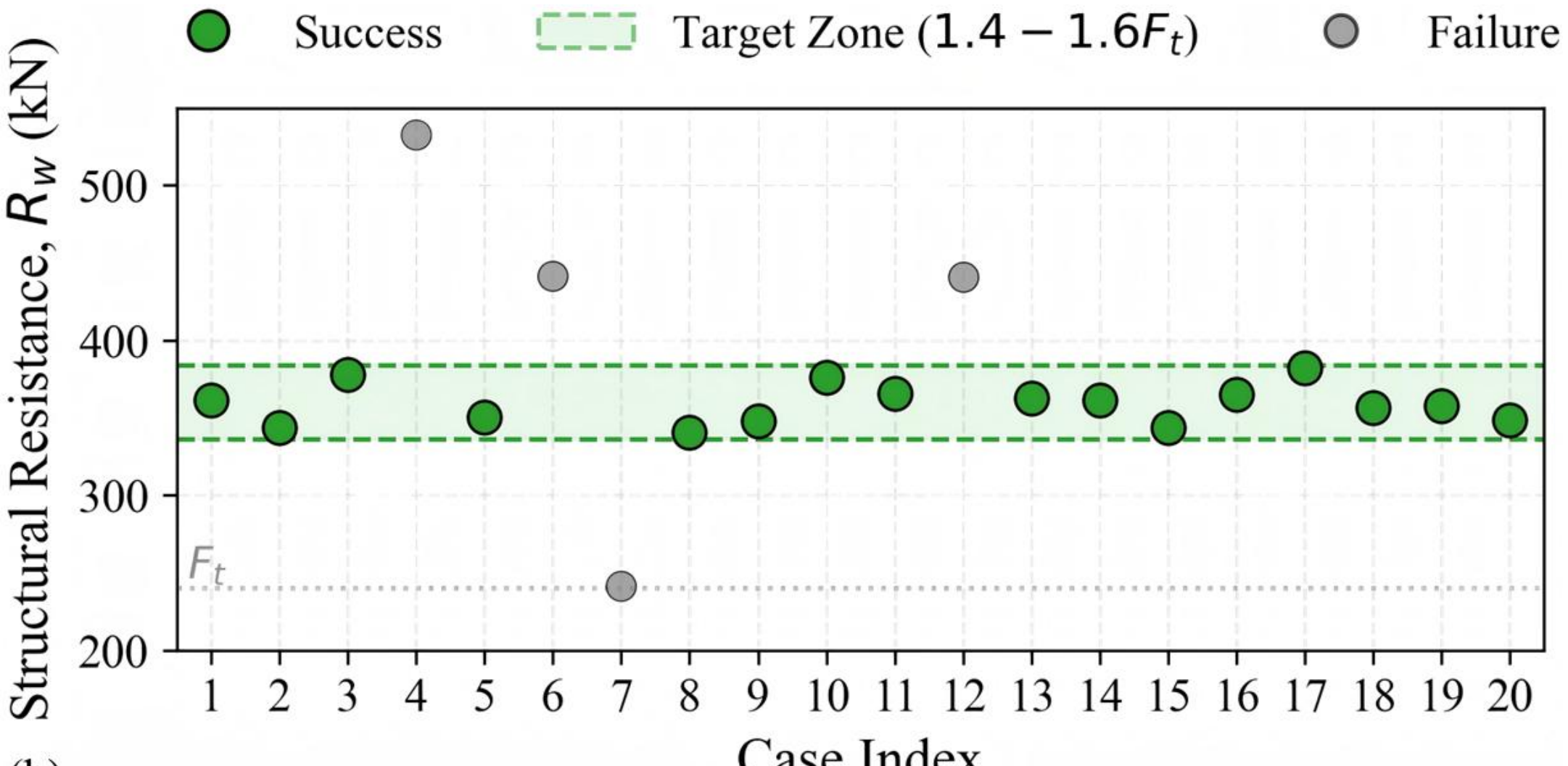

(b)

Figure A2. Target-interval compliance results of TL- 3 barriers using multi-agent framework: (a) MAF-DS-8B; (b) MAF-DS-32B.

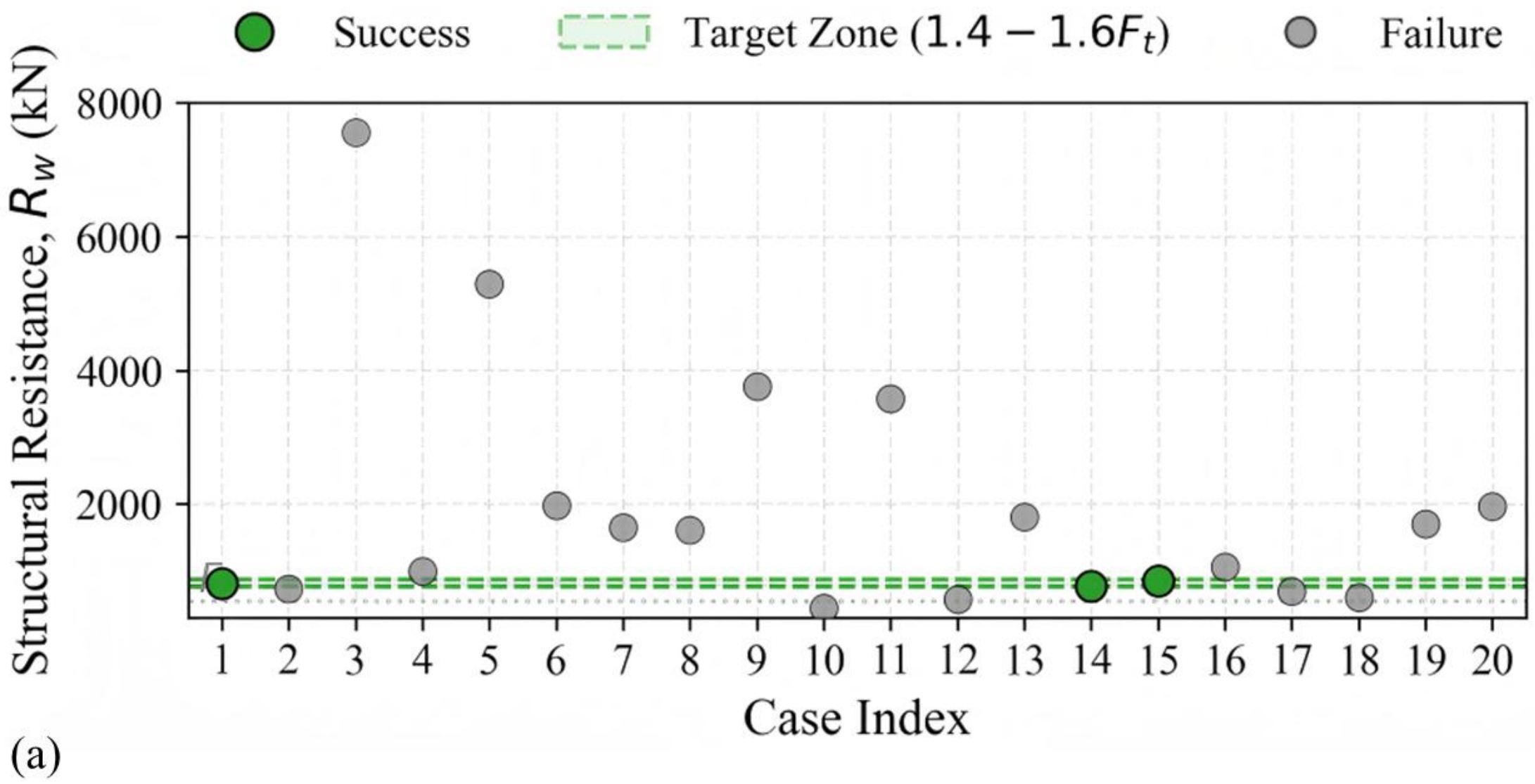

(a)

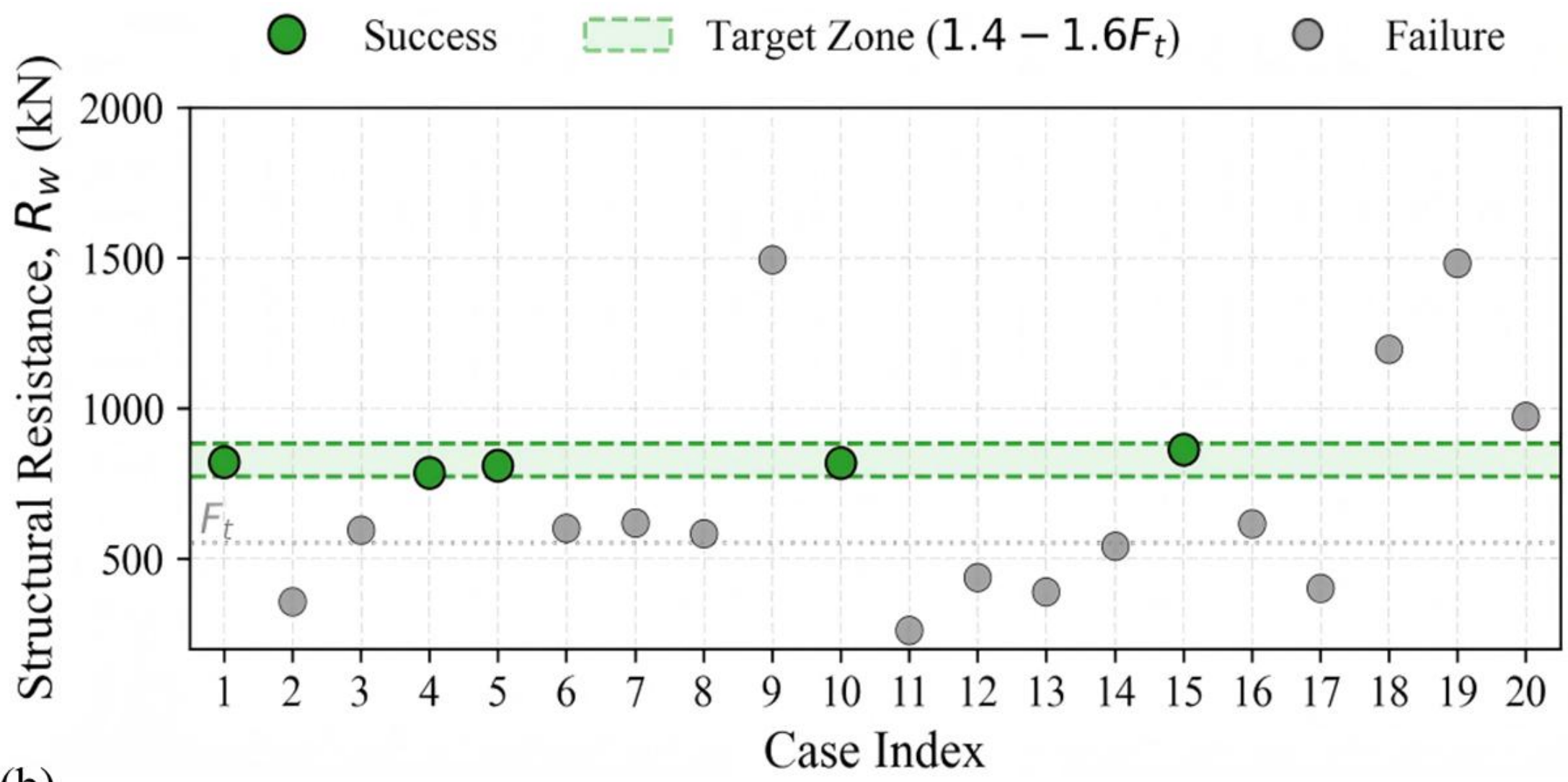

(b)

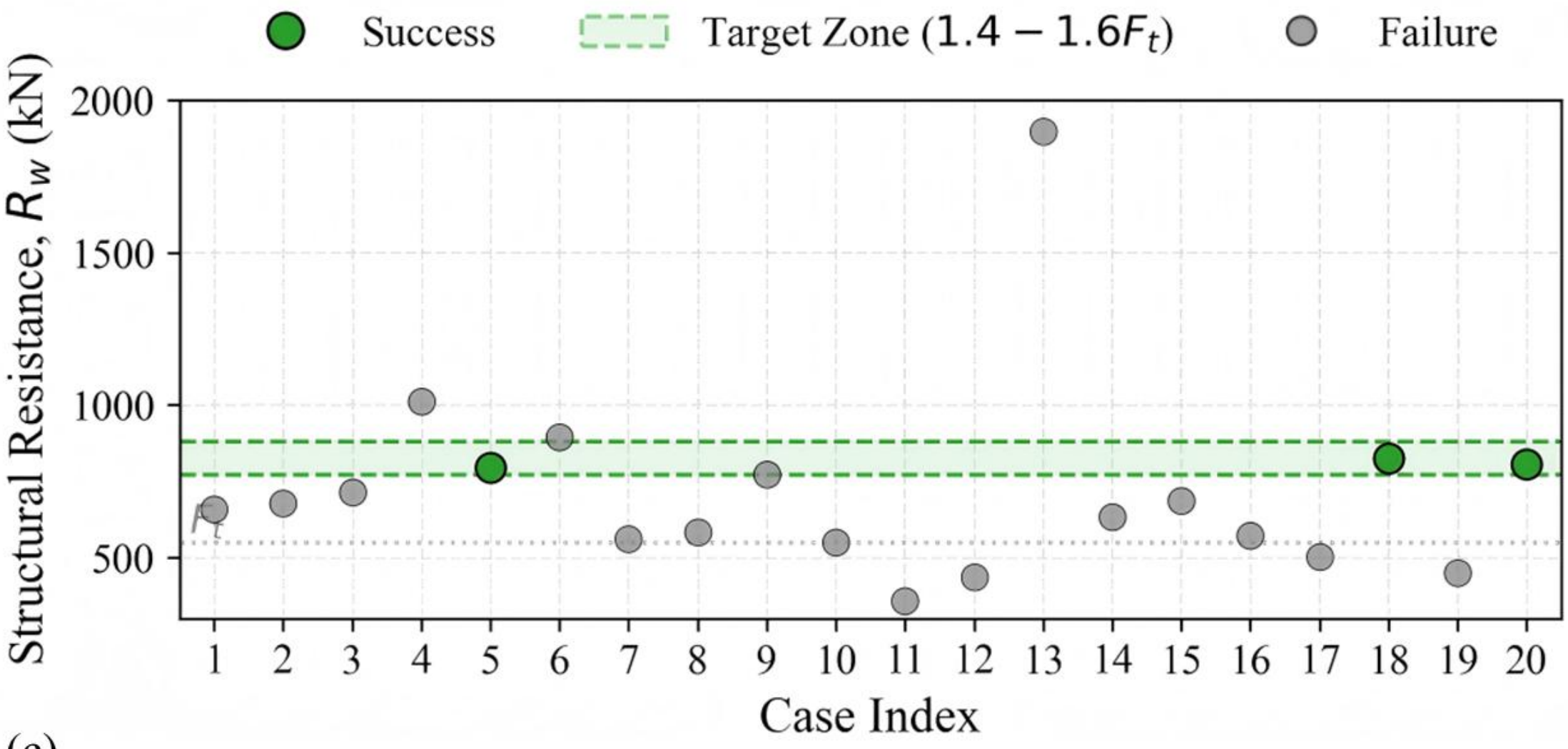

(c)

Figure A3. Target-interval compliance results of TL- 5 barriers using general large language models: (a) DS-8B; (b) DS-32B; (c) DS-671B

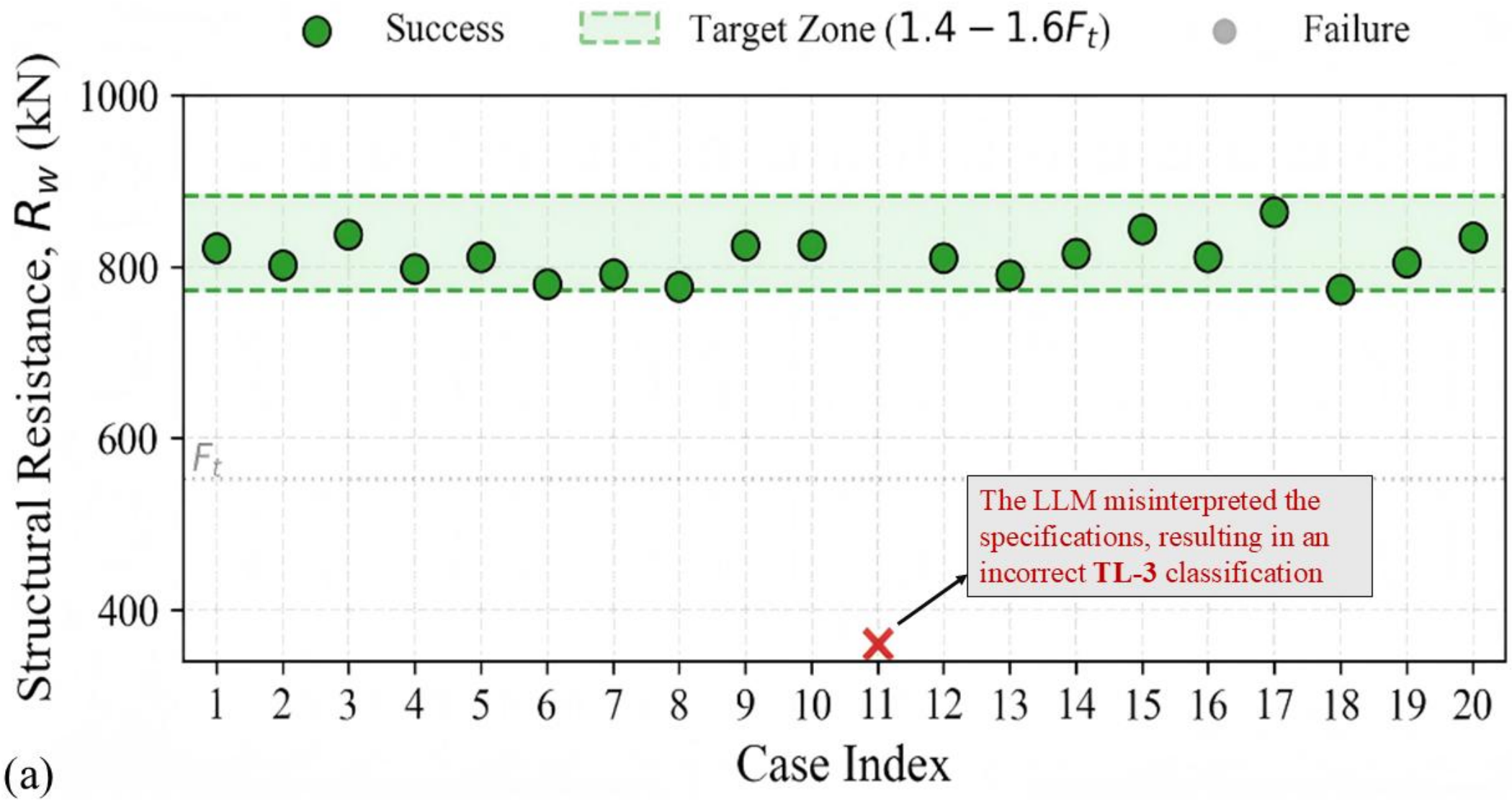

(a)

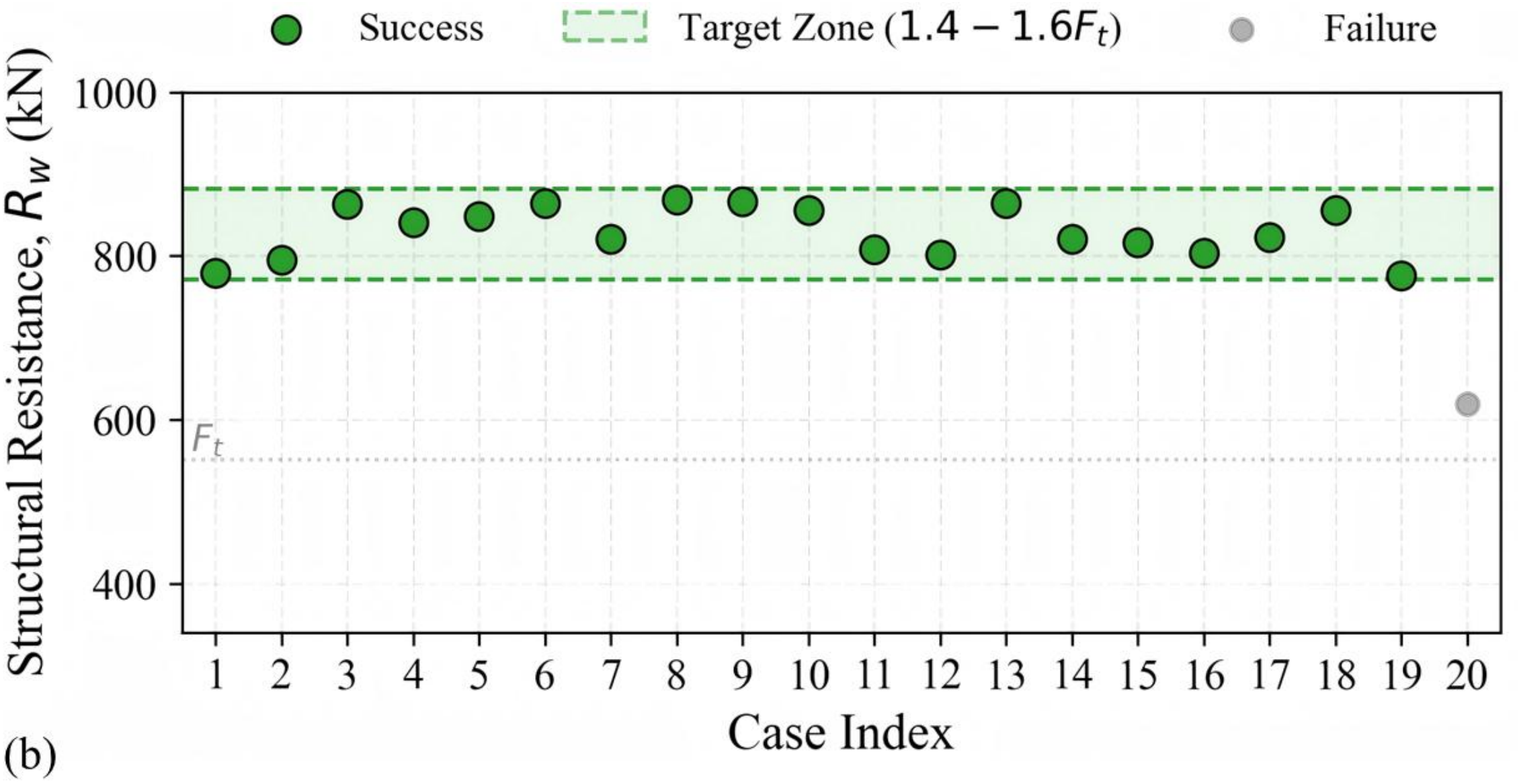

(b)

Figure A4. Target-interval compliance results of TL- 5 barriers using multi-agent framework: (a) MAF-DS-8B; (b) MAF-DS-32B.

Table A1. Design Resistance and Force Ratios for NCHRP 1109 TL-Series Bridge Rail Design

| Case | Barrier Type | $R_w$ (kips) | $F_t$ (kips) | $R_w/F_t$ |
|---|---|---|---|---|
| 1 | TL-5 Concrete Barrier | 397 | 162 | 2.45 |
| 2 | TL-5 Concrete Barrier | 258 | 162 | 1.59 |
| 3 | TL-4 Steel Post-and-Beam Railing | 79.04 | 54 | 1.46 |
| 4 | TL-4 Railing | 119.64 | 68 | 1.76 |
| 5 | TL-4 Steel Post-and-Beam Railing | 79.04 | 54 | 1.46 |
| 6 | TL-4 Railing | 119.64 | 68 | 1.76 |
| 7 | TL-3 Curb-Mounted Steel Post and Aluminum Rails | 57.98 | 54 | 1.07 |
| 8 | TL-3 Curb-Mounted Steel Post and Aluminum Rails | 110.51 | 70 | 1.58 |
| 9 | TL-3 Curb-Mounted Steel Post and Aluminum Rails | 86.07 | 70 | 1.23 |
| | Average | —— | —— | 1.49 |

Table A2. Parameter fields, fallback values, and deterministic correction rules used in the implemented workflow

| Parameter | Data type | Unit | Fallback value or programmed action |
|---|---|---|---|
| grade | String | - | TL-3 |
| $H$ | Float | mm | 813.0 |
| $B_{top}$ | Float | mm | 230.0 |
| $B_{bottom}$ | Float | mm | 400.0 |
| $c$ | Float | mm | 38.0 |
| $f_c$ | Float | Mpa | 28.0 |
| $f_{y1}$ | Float | Mpa | 400.0 |
| $f_{y2}$ | Float | Mpa | 400.0 |
| $s$ | Float | mm | 150.0 |
| $n$ | Integer | - | 8 |
| $d_z$ | Float | mm | $d_z$ to 16 mm |
| $d_k$ | Float | mm | $d_k$ to 10 mm |

The listed fallback values are programmatic completion values used when a field cannot be extracted; they are not presented as universally recommended design values. The final two rows report the fixed correction rules applied to anomalous reinforcement diameters.